\documentclass[letterpaper]{article}

\usepackage{natbib,alifeconf, subcaption, graphicx}  

\title{Normalisation of Weights and Firing Rates in Spiking Neural Networks with Spike-Timing-Dependent Plasticity}
\author{Katarzyna Kozdon$^{1}$, Peter Bentley$^{1,2}$ \\
\mbox{}\\
$^1$University College London, London, United Kingdom \\
$^2$Braintree Ltd, London, United Kingdom \\
k.kozdon@cs.ucl.ac.uk} 

\begin{document}
\maketitle

\begin{abstract}
Maintaining the ability to fire sparsely is crucial for information encoding in neural networks. Additionally, spiking homeostasis is vital for spiking neural networks with changing numbers of weights and neurons. We discuss a range of network stabilisation approaches, inspired by homeostatic synaptic plasticity mechanisms reported in the brain. These include weight scaling, and weight change as a function of the network's spiking activity. We tested normalisation of the sum of weights for all neurons, and by neuron type. We examined how this approach affects firing rate and performance on clustering of time-series data in the form of moving geometric shapes. We found that neuron type-specific normalisation is a promising approach for preventing weight drift in spiking neural networks, thus enabling longer training cycles. It can be adapted for networks with architectural plasticity.    
\end{abstract}

\section{Introduction}
Learning systems need to maintain a balance between plasticity and stability. In the brain, mechanisms such as long-term potentiation (LTP) and long-term depression (LTD) are vital for learning, but on their own they would lead to a powerful destabilisation of neuronal activity (\cite{Abbott2000, Miller1994}). Strengthening of the synapses would lead to the increased likelihood of postsynaptic firing, whereas weakening of the synapses would lead to their silencing and eventually a cascade of further downstream changes in the same direction.\par
Similar problems are encountered in artificial intelligence. In  unsupervised spiking neural networks (SNN) with spike-timing-dependent plasticity (STDP), weight drift is a known problem which impedes learning by leading to either weight silencing or saturation (\cite{Miller1994, Abbott2000, Oja1982}). In these networks, weight update is a function of the spike timing in a pair of connected neurons. Unlike in non-spiking supervised neural networks (NNN), the weight update is not aimed to minimise the loss function. Techniques such as L1 (\cite{RobertTibshirani1997}) and L2 (\cite{Nigam}) regularisation used to prevent excessive weight growth and parameter overfitting in NNN are not directly applicable to SNN, because of the complex relationship between the weights and spiking activity. \par
Additional difficulties arise when creating networks with plastic architecture. What value should be assigned to new weights, associated either with new synapses of existing neurons, or new neurons? The selected weight will not only affect the capacity of the new connection to learn, but will also cause downstream effects which are hard to predict in a complex systems, and may not be desirable. Thus, in order to utilise SNN for data processing, we need to identify methods preventing weight drift, silencing and saturation of spiking. Ideally, these methods should be applicable to SNN with changing numbers of neurons and synapses. These normalisation mechanisms cannot simply counteract the changes as it would effectively lead to erasing memories and undoing learning (\cite{Fox2017}).\par  
One of the classical ways to tame Hebbian learning--induced bistability is the Bienenstock, Cooper and Munro model (\cite{Bienenstock1982}), according to which correlated spiking leads to LTP if the postsynaptic rate is higher than threshold, and to LTD if the firing rate is lower. The threshold changes as a function of average postsynaptic firing rate. However, there is no direct evidence that this mechanism exists in the brain.\par 
Oja rule (\cite{Oja1982}) is another classical approach to stabilising networks with STDP. It extends Hebbian plasticity with scaling down synaptic efficacy as a function of the square of the firing rate.\par 
Overall, models suggest that weight normalisation mechanisms are crucial for preserving the functionality of SNN. However, little research exploring and comparing different weight normalisation methods in applied SNN, and in SNN with plastic architecture, has been published.\par
 The brain has developed homeostatic mechanisms for preserving its functionality while changing during development and learning. These mechanisms act both at the global (\cite{Turrigiano2008}) and local (\cite{Yu2009}) level. Four main forms have been distinguished (\cite{Fox2017}):
\begin{itemize}
    \item synaptic scaling (SS), 
    \item adjustment of the levels of inhibition and excitation,
    \item basal firing rate homeostasis,
    \item plasticity of the neuronal membrane properties.
\end{itemize}\par
 Relevant to our project, it has been hypothesised that SS may be needed ''to ensure that firing rates do not become saturated during developmental changes in the number and strength of synaptic inputs, as well as stabilising synaptic strengths during Hebbian modification and facilitating competition between synapses" (\cite{Turrigiano1998}). Another attractive property of SS is stabilisation of neuronal activity without disrupting information storage and processing related to differences in synaptic strengths.\par  

\section{Model overview and experimental setup}
The model set-up is summarised in table \ref{table:norm_setup}, and is a modification of \cite{Kozdon2018}. Briefly, we used the exponential leaky integrate-and-fire model (\cite{Barranca2014, Fourcaud-Trocme2003}). Our network was composed of three feed-forward, fully-connected layers with 500, 50 and 10 neurons. 80\% of the neurons were excitatory and 20\% were inhibitory.\par
\begin{figure}[!t]
\centering
\includegraphics[width=0.45\textwidth, keepaspectratio]{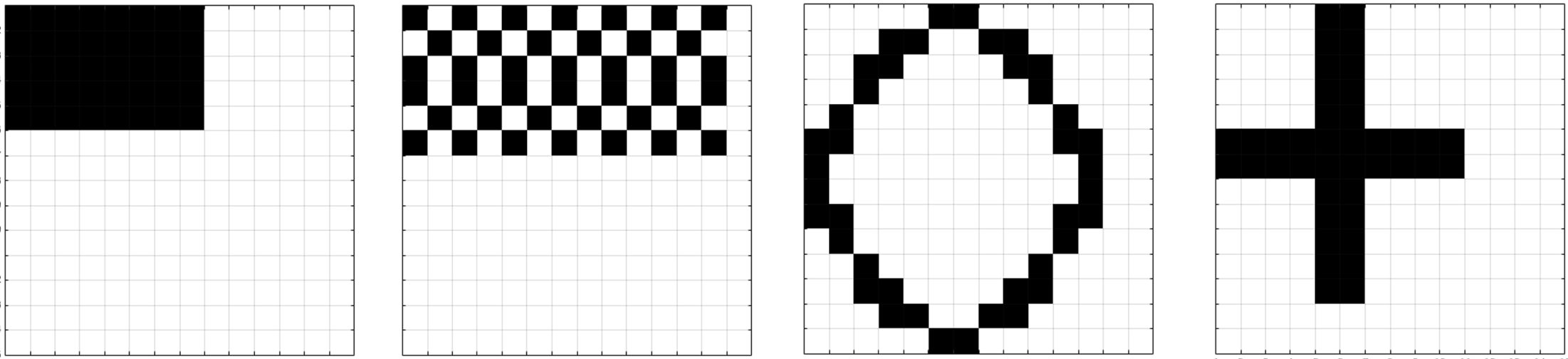}
\caption[Examples of bitmaps used as input data.]
{Examples of bitmaps used as input data: a square, grid, ellipse and cross, each composed of 40 black pixels. One of the shapes was placed at a random location within the visual field (20 x 25 pixels) and moved up, down, left or right.}
\label{fig:EAInputs}
\end{figure}

Our input data were bitmaps representing four different shapes (Fig. \ref{fig:EAInputs}) moving in one of four directions at the speed of one pixel per frame. Each bitmap had 500 pixels, and each iteration of the network activity corresponded to a new frame.\par
We used evolutionary approach described in detail in \cite{Kozdon2018}. Clustering inputs by movement direction was used as a measure of fitness. A population of networks was initialised with random starting weights and learning hyperparameters. After training and testing, the top third of the networks were treated as non-sexually reproducing parents, each giving rise to three children: a child-clone, a child-clone which undergoes another training cycle, and a child with a mutation in one of the four learning parameters and exposed to another cycle of training.\par
We examined the spiking activity in each layer, and performance of the networks.\par

\begin{table}[!t]
\caption[Model hyperparameters]{Model hyperparameters.}
\label{table:norm_setup}
\begin{center}
\resizebox{0.4\textwidth}{!}{\begin{tabular}{l l l}
\hline
\multicolumn{3}{l}{\textbf{Neural networks}}\\
\hline
input neurons & 500\\
hidden neurons & 50\\
output neurons & 10\\
\hline
\multicolumn{3}{l}{\textbf{Training and testing set-up}}\\
\hline
frames per training input & 10\\
inputs per training cycle & 20\\
frames per testing input & 50\\
inputs per testing cycle & 80\\
time step & 0.02 s\\
\hline
\multicolumn{3}{l}{\textbf{Electrophysiological parameters}}\\
\hline
& \emph{excitatory} & \emph{inhibitory}\\
resting potential & -63.70 & -59.3\\
reset potential &   -68.7 & -64.3 \\
firing threshold & -45.80 & -33.3\\
resistance & 67.70 & 133.3\\
$\tau$ & 16.70 & 36.5\\
\hline
\multicolumn{3}{l}{\textbf{Evolution constraints}}\\
\hline
& \emph{min} & \emph{max} \\
\cline{2-3}
LTP & 0.001 & 0.1\\
inhLTP & 0.001 & 0.1\\
LTD & 0.001 & 0.1\\
discharge & 0.06 & 5.0\\
\end{tabular}}
\end{center}
\end{table}

\subsection{Experimental setup: global weight normalisation}
Normalisation was performed after each training epoch (10 frames of the input data). Based on our preliminary experiments, we selected the normalised sum of weights $w_{norm}$ to be tied to the number of all neurons ($n_n$) in the network and equal:
\begin{equation}
w_{norm}= n_n \cdot 100
\label{eq:normalisedSumOfWeights}
\end{equation}
We tested six set-ups:
\begin{enumerate}
    \item No normalisation. The weights were capped at 4 as in the previous paper (\cite{Kozdon2018}) (``control\_capped").
    \item No normalisation and weight-capping (``control").
    \item Normalisation with weight-capping (``norm\_capped"). 
    \item Normalisation without weight-capping (``norm").
    \item Separate normalisation for excitatory and inhibitory neurons, with weight-capping (``norm\_capped\_ie"). 
    \item Separate normalisation for excitatory and inhibitory neurons, without weight-capping. A proportion of $w_{norm}$ was assigned to each neuronal population; 80\% was assigned to excitatory and 20\% to inhibitory synapses (``norm\_ie"). 
\end{enumerate}

We analysed a population of 12 networks over 20 generations. 10 repeats were performed. Significance was calculated using the Kruskal-Wallis test.\par

\section{Results}
\subsection{Global weight normalisation}
\begin{figure}[!ht]
\centering
\begin{subfigure}{0.45\textwidth}
\includegraphics[width=\textwidth]{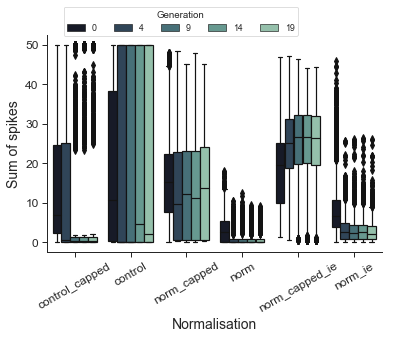}
\subcaption
{Hidden layer.}
\end{subfigure}
\begin{subfigure}{0.45\textwidth}
\includegraphics[width=\textwidth]{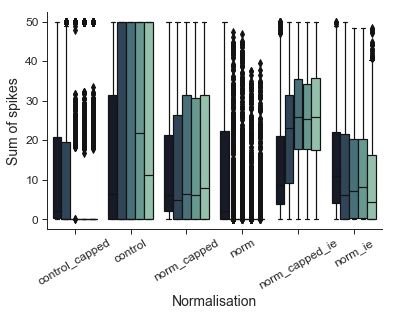}
\subcaption
{Output layer.}
\end{subfigure}
\caption[Analysis of spiking activity.]
{Analysis of spiking activity. Median number of spikes per neuron per 50 iterations. Bars indicated standard deviation, and dots outliers.}
\label{fig:spiking}
\end{figure}

Effects of normalisation were different depending on the presence of the weight cap \ref{fig:spiking}).\par
In networks without normalisation but with weight-capping (``control\_capped"), spiking in the hidden and output layers (Fig. \ref{fig:spiking} a and b) decreased with time. Networks without normalisation and weight-capping (``control") demonstrated a wide range of spiking patterns, and ranged from fully saturated to fully silent. Both of these un-normalised approached lead to activity patterns which are undesirable for information processing.\par
Networks with weight caps and normalisation (``norm\_capped" and ``norm\_capped\_ie") demonstrated the most stable spiking behaviour, and were resistant to silencing while avoiding saturation during the examined period.\par
\begin{figure}[!t]
\centering
\includegraphics[width=0.45\textwidth, keepaspectratio]{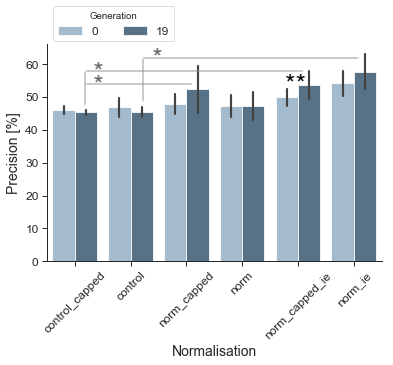}
\caption[Fitness of networks with different normalisation methods.]
{Fitness of the networks. Double asterisk indicates significant difference between generation 0 and 19. Single asterisks indicate significant precision differences between normalised networks and their respective controls in generation 19. Bars indicate standard deviation.}
\label{fig:performance}
\end{figure}

 
Networks with normalisation and capping (``norm\_capped"), and both set-ups with neuron type-specific weight normalisation (``norm\_capped\_ie" and ``norm\_ie") were significantly more precise than the controls (Fig. \ref{fig:performance}). Their performance also improved with time; normalisation with weight-capping had a lower precision than normalisation without the cap, but a steeper learning rate.\par 

\section{Discussion}
Our results demonstrate that even simple weight normalisation approaches can reduce saturation and silencing in SNN. These effects translated into improved performance on the unsupervised clustering tasks.\par
The simplest approach to network stabilisation is a global normalisation of the sum of weights, irrespective of the neuron type. The sum of weights is a function of the number of neurons, which gives it a potential to be suitable for networks with a changing number of neurons.\par 
We also tested a neuron type-dependent modification of this approach, and normalised the sum of excitatory and inhibitory weights separately, assigning a proportion of the sum of weights to each population. This approach aimed to preserve the balance between the two. However, as the activity of each neuron depends on the strengths and types of all incoming weights, this is an indirect and global way of balancing excitation and inhibition, and does not normalise the exact balance between these two. This approach could be further extended to networks with multiple neuronal populations, even if the numbers of neurons fluctuate.\par 
We are currently testing other approaches:
\begin{itemize}
    \item Normalisation of the spiking activity level:
    \begin{itemize}
        \item of the output neurons.
        \item of all neurons.
    \end{itemize}
    \item Activity-based adjustment of the level of inhibition and excitation. 
\end{itemize}

When stabilising SNN, we primarily care about preserving their ability to encode information using spikes. Weight normalisation is an indirect approach to achieve this; a more direct approach would be to adjust the weights based on the spiking activity of the network. Spiking of the output layer is our direct readout. It is desired for the output neurons to fire sparsely (both activation of all neurons and total silence are to be avoided) so that multiple classes of input data can be encoded. Thus, we could focus solely on adjusting weights as a function of spiking in the output layer. This approach benefits from direct focus on stabilising our readout. However, spiking of the output layer may not provide enough resolution to fine-tune weights of the whole network.  Alternatively, as the activity of the output layer is a function of the activity of the preceding layers, we could change the weights as a function of spiking of the whole network. This approach is less direct, but the breadth of observed behaviours may allow us to fine-tune the activity of the network.\par
Weight update can be multiplicative or additive. In the first case, weights are changed proportionally to their size. This method affects distribution of the weights. In additive weight update, a set amount $\Delta w$ is added or subtracted from each weight. Weights should not be allowed to drop below 0 to avoid switching between excitation and inhibition. This method allows to activate silent synapses.\par  

Selective scaling of inhibitory and excitatory neurons as the function of spiking activity of the whole network or the output layer is a possible modification of this approach. If the number of spikes exceeds the permitted range, excitatory weights can be decreased and/or inhibitory increased; if spiking is below the permitted range, the reverse is executed. This approach can be adapted to minimise the weights without silencing of the network, and thus can be a form of activity-dependent mechanism reducing overfitting.\par

\section{Conclusions}
Spiking neural networks with STDP require a mechanism for preventing weight drift. Our results show that even simple normalisation of the sum of weights, tied to the number of neurons provides stabilisation. Interestingly, the results are improved when normalisation was neuron type-specific, maintaining the sum of inhibitory and excitatory weights separately. This method is beneficial for maintaining trainable weights in networks with static architecture; it can also be implemented in networks with fluctuating number of neurons and weights, and with multiple neuronal populations, thus enabling their training and stabilisation of their activity levels.

\section{Acknowledgements}

This work was supported by Biotechnology and Biological Sciences Research Council and the London Interdisciplinary Biosciences Consortium.

\footnotesize
\bibliographystyle{apalike}
\bibliography{references} 

\begin{thebibliography}{}

\bibitem[Abbott and Nelson, 2000]{Abbott2000}
Abbott, L.~F. and Nelson, S.~B. (2000).
\newblock {Synaptic plasticity: Taming the beast}.
\newblock {\em Nature Neuroscience}, 3(11s):1178--1183.

\bibitem[Barranca et~al., 2014]{Barranca2014}
Barranca, V.~J., Johnson, D.~C., Moyher, J.~L., Sauppe, J.~P., Shkarayev,
  M.~S., Kova{\v{c}}i{\v{c}}, G., and Cai, D. (2014).
\newblock {Dynamics of the exponential integrate-and-fire model with slow
  currents and adaptation.}
\newblock {\em Journal of computational neuroscience}, 37(1):161--80.

\bibitem[Bienenstock et~al., 1982]{Bienenstock1982}
Bienenstock, E.~L., Cooper, L.~N., and Munro, P.~W. (1982).
\newblock {Theory for the development of neuron selectivity: orientation
  specificity and binocular interaction in visual cortex.}
\newblock {\em The Journal of neuroscience : the official journal of the
  Society for Neuroscience}, 2(1):32--48.

\bibitem[Fourcaud-Trocme et~al., 2003]{Fourcaud-Trocme2003}
Fourcaud-Trocme, N., Hansel, D., Vreeswijk, C.~V., and Brunel, N. (2003).
\newblock {How spike mechanisms determine response to fluctuating inputs}.
\newblock {\em The Journal of Neuroscience}, 23(37):11628--11640.

\bibitem[Fox and Stryker, 2017]{Fox2017}
Fox, K. and Stryker, M. (2017).
\newblock {Integrating Hebbian and homeostatic plasticity: Introduction}.
\newblock {\em Philosophical Transactions of the Royal Society B: Biological
  Sciences}, 372(1715).

\bibitem[Kozdon and Bentley, 2018]{Kozdon2018}
Kozdon, K. and Bentley, P. (2018).
\newblock {The Evolution of Training Parameters for Spiking Neural Networks
  with Hebbian Learning}.
\newblock {\em The 2018 Conference on Artificial Life}, pages 276--283.

\bibitem[Miller and Mackay, 1994]{Miller1994}
Miller, K.~D. and Mackay, D. J.~C. (1994).
\newblock {The Role of Constraints in Hebbian Learning}.
\newblock {\em Neural computation}, 6:100--126.

\bibitem[Nigam et~al., 1999]{Nigam}
Nigam, K., Lafferty, J., and Mccallum, A. (1999).
\newblock {Using Maximum Entropy for Text Classification}.
\newblock {\em IJCAI-99 Workshop on Machine Learning for Information
  Filtering}, pages p. 61-- 67.

\bibitem[Oja, 1982]{Oja1982}
Oja, E. (1982).
\newblock {Simplified neuron model as a principal component analyzer}.
\newblock {\em Journal of Mathematical Biology}, 15(3):267--273.

\bibitem[Tibshirani, 1997]{RobertTibshirani1997}
Tibshirani, R. (1997).
\newblock {The LASSO method for variable selection in the Cox model}.
\newblock {\em Statistics in Medicine}, 16(4):385--395.

\bibitem[Turrigiano, 2008]{Turrigiano2008}
Turrigiano, G.~G. (2008).
\newblock {The Self-Tuning Neuron: Synaptic Scaling of Excitatory Synapses}.

\bibitem[Turrigiano et~al., 1998]{Turrigiano1998}
Turrigiano, G.~G., Leslie, K.~R., Desai, N.~S., Rutherford, L.~C., and Nelson,
  S.~B. (1998).
\newblock {Activity-dependent scaling of quantal amplitude in neocortical
  neurons}.
\newblock {\em Nature}, 391(6670):892--896.

\bibitem[Yu and Goda, 2009]{Yu2009}
Yu, L.~M. and Goda, Y. (2009).
\newblock {Dendritic signalling and homeostatic adaptation}.

\end{thebibliography}

\end{document}